\documentclass{article}
\usepackage{spconf,amsmath,graphicx}

\usepackage{color}
\usepackage{url}  
\usepackage{graphicx}  
\usepackage{amsmath}
\usepackage{amssymb}
 \usepackage{booktabs}
 \usepackage{multirow}
\newcommand{\tabincell}[2]{\begin{tabular}{@{}#1@{}}#2\end{tabular}}

\title{Frame Attention Networks for Facial Expression Recognition in Videos}
\name{Debin Meng, Xiaojiang Peng$^*$,  Kai Wang,  Yu Qiao
\thanks{$^*$Xiaojiang Peng is the corresponding author. Email: xj.peng@siat.ac.cn}
\thanks{This work was supported by the National Natural Science Foundation of China (U1613211, U1713208), Shenzhen Research Program (JCYJ20170818164704758, JSGG20180507182100698), and International Partnership Program of Chinese Academy of Sciences (172644KYSB20150019).}}
\address{Shenzhen Institutes of Advanced Technology, Chinese Academy of Science, Shenzhen, China\\ Shenzhen Key Lab of Computer Vision and Pattern Recognition, Shenzhen, China \\ University of Chinese Academy of Sciences, Beijing, China
\\ michaeldbmeng19@outlook.com, \{xj.peng, kai.wang, yu.qiao\}@siat.ac.cn}

\usepackage{fancyhdr}

\begin{document}
%
\maketitle

\thispagestyle{fancy}
\fancyhead{}
\lhead{}
\lfoot{\footnotesize{Copyright 2019 IEEE. Published in the IEEE 2019 International Conference on Image Processing (ICIP 2019), scheduled for 22-25 September 2019 in Taipei, Taiwan. Personal use of this material is permitted. However, permission to reprint/republish this material for advertising or promotional purposes or for creating new collective works for resale or redistribution to servers or lists, or to reuse any copyrighted component of this work in other works, must be obtained from the IEEE. Contact: Manager, Copyrights and Permissions / IEEE Service Center / 445 Hoes Lane / P.O. Box 1331 / Piscataway, NJ 08855-1331, USA. Telephone: + Intl. 908-562-3966.}}
\cfoot{}
\rfoot{}

\copyrightnotice{
\copyright\ IEEE 2019}\toappear{To appear in {\it Proc.\ ICIP2019, September 22-25, 2019, Taipei, Taiwan}}

\begin{abstract}
The video-based facial expression recognition aims to classify a given video into several basic emotions. How to integrate facial features of individual frames is crucial for this task. In this paper, we propose the Frame Attention Networks (FAN)\footnote{Code is available at https://github.com/Open-Debin/Emotion-FAN}, to automatically highlight some discriminative frames in an end-to-end framework.
The network takes a video with a variable number of face images as its input and produces a fixed-dimension representation. The whole network is composed of two modules. The feature embedding module is a deep Convolutional Neural Network (CNN) which embeds face images into feature vectors. The frame attention module learns multiple attention weights which are used to adaptively aggregate the feature vectors to form a single discriminative video representation.
We conduct extensive experiments on CK+ and AFEW8.0 datasets. Our proposed FAN shows superior performance compared to other CNN based methods and achieves state-of-the-art performance on CK+. 

\end{abstract}
\section{Introduction}
\label{sec:intro}

Automatic facial expression recognition (FER) has recently attracted increasing attention in academia and industry due to its wide range of applications such as affective computing, intelligent environments, and multimodal human-computer interface (HCI). 
Though great progress have been made recently, facial expression recognition in the wild remains a challenging problem due to
large head pose, illumination variance, occlusion, motion blur, etc.

Video-based facial expression recognition aims to classify a video into several basic emotions, such as happy, angry, disgust, fear, sad, neutral, and surprise. Given a video, the popular FER pipeline with a visual clue (FER with an audio clue is out of the scope of this paper) mainly includes three steps, namely frame preprocessing,  feature extraction, and classification. Especially, frame preprocessing refers to face detection, alignment, illumination normalizing and so on. Feature extraction or video representation is the key part for FER which encodes frames or sequences into compact feature vectors. These feature vectors are subsequently fed into a classifier for prediction.

Feature extraction methods for video-based FER can be roughly divided into three types: static-based methods, spatial-temporal methods, and geometry-based methods.

Static-based feature extraction methods mainly inherit those methods from static image emotion recognition which can be both hand-crafted~\cite{Littlewort2006Dynamics,Shan2009Facial} and learned~\cite{Tang2013Deep,Bargal2016Emotion,Knyazev2017Convolutional}. For the hand-crafted features, Littlewort \textit{et al}.~\cite{Littlewort2006Dynamics} propose to use a bank of 2D Gabor filters to extract facial features for video-based FER. Shan \textit{et al}.~\cite{Shan2009Facial} use local binary patterns (LBP) and LBP histogram for facial feature extraction. 
For the learned features, 
Tang~\cite{Tang2013Deep} utilizes deep CNNs for feature extraction, and win the FER2013.
Some winners in audio-video emotion recognition task of EmotiW2016 and EmotiW2017 only use static facial features from deep CNNs trained on large face datasets or trained with multi-level supervision~\cite{Bargal2016Emotion,Knyazev2017Convolutional}. 

Spatial-temporal methods aim to model the temporal or motion information in videos. The  Long Short-Term Memory (LSTM)~\cite{Hochreiter1997Long}, and C3D~\cite{tran2015learning} are two widely-used spatial-temporal methods for video-based FER.  LSTM derives information from sequences by exploiting the fact that feature vectors are connected semantically for successive data. This pipeline is widely-used in the EmotiW challenge, e.g. \cite{Liu2016Video,ouyang2017audio,Vielzeuf2017Temporal,Yan2018Multi}. C3D, which is originally developed for video action recognition, is also popular in the EmotiW challenge. 

 Geometry based methods~\cite{Jung2015Joint,Yan2018Multi} aim to model the motions of key points in faces which only leverage the geometry locations of facial landmarks in every video frames. Jung \textit{et al}.~\cite{Jung2015Joint} propose a deep temporal appearance-geometry network (DTAGN) which first alternately concatenates the x-coordinates and y-coordinates of the facial landmark points from each frame after normalization and then concatenates these normalized points over time for a one-dimensional trajectory signal of each sequence. Yan \textit{et al}.~\cite{Yan2018Multi} construct an image-like map by stretching all the normalized facial point trajectories in a sequence together as the input of a CNN. 
 
 Among all the above methods, static-based methods are superior to the others according to several winner solutions in EmotiW challenges.  
To obtain a video-level result with varied frames, a frame aggregation operation is necessary for static-based methods. For frame aggregation, Kahou \textit{et al}.~\cite{Kahou2013Combining} concatenate the \textit{n}-class probability vectors of 10 segments to form a fixed-length video representation by frame averaging or frame expansion. 
Bargal \textit{et al}.~\cite{Bargal2016Emotion} propose a statistical encoding module (STAT) to aggregate frame features which compute the mean, variance, minimum, and maximum of the frame feature vectors. 

One limitation of these existing aggregation methods is that they ignore the importance of frames for FER. For example, some faces in Figure \ref{fig_fan} are representative for the `happy' category while the others not.
In this paper, inspired by the attention mechanism~\cite{vaswani2017attention} of machine translation and the neural aggregation networks~\cite{yang2017neural} of video face recognition, we propose the Frame Attention Networks (FAN) to adaptively aggregate frame features.
The FAN is designed to learn self-attention kernels and relation-attention kernels for frame importance reasoning in an end-to-end fashion. The self-attention kernels are directly learned from frame features while the relation-attention kernels are learned from the concatenated features of a video-level anchor feature and frame features.
We conduct extensive experiments on CK+ and AFEW8.0 (EmotiW2018) datasets. Our proposed FAN shows superior performance compared to other CNN based methods with only facial features and achieves state-of-the-art performance on CK+.

\section{Frame Attention Networks}
\label{sec:Frame Attention Networks}

\begin{figure}[htp]
\centering
\includegraphics[width=0.95\linewidth]{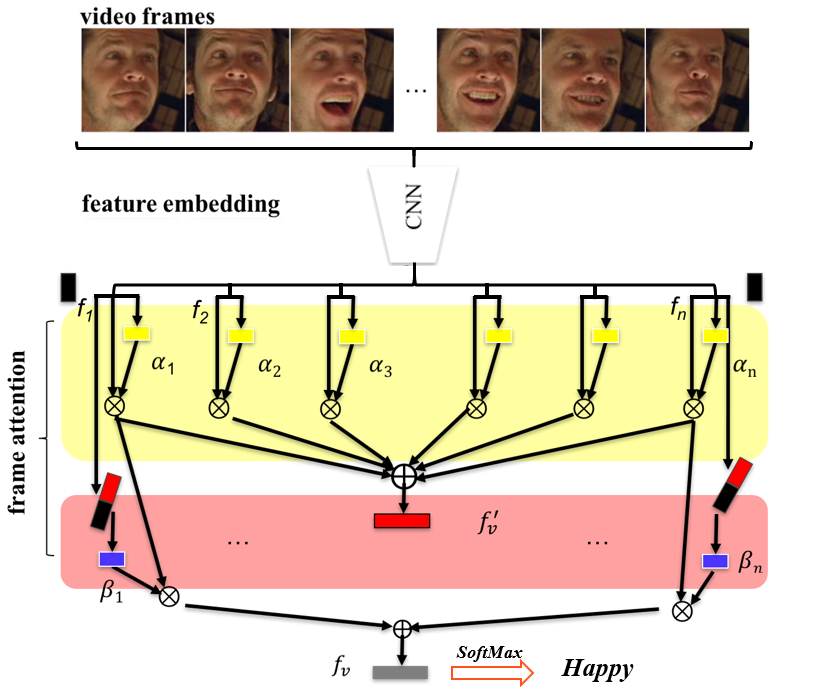}
\caption{Our proposed frame attention network architecture.}
\label{fig_fan}
\end{figure}

We propose Frame Attention Networks (FAN) for video-based facial expression recognition (FER). Figure \ref{fig_fan} illustrates the framework of our proposed FAN. It takes a facial video with a variable number of face images as its input and produces a fixed-dimension feature representation for FER. The whole network consists of two modules: feature embedding module and frame attention module. The feature embedding module is a deep CNN which embeds each face image into a feature vector. The frame attention module learns two-level attention weights, i.e. \textit{self-attention weights} and \textit{relation-attention weights}, which are used to adaptively aggregate the feature vectors to form a single discriminative video representation. 

Formally, we denote a video with $n$ frames as $\mathbf{V}$, and its frames as ${I_1},{I_2}, \cdots ,{I_n}$,  and the facial frame features are \{$f_1,\cdots, {f_n}$\}.

\textbf{Self-attention weights}. With individual frame features, our FAN first applies a FC layer and a sigmoid function to assign coarse self-attention weights. Mathematically, the self-attention weight of the $i$-th frame is defined by: 

\begin{equation}
\label{eq:alpha}
\alpha_{i} = \sigma(f_i^T\mathbf{q}^0)
\end{equation}
where $\mathbf{q}^0$ is the parameter of FC, $\sigma$ denotes the sigmoid function. 
With these self-attention weights, we aggregate all the input frame features into a global representation $f'_v$ as follows,
\begin{equation}
f'_v= \frac {\sum_{i=1}^{n}\alpha_{i}f_{i}}{\sum_{i=1}^{n}\alpha_{i}}.
\end{equation}

We use $f'_v$ as a video-level global anchor for learning further accurate relation-attention weights. 

\textbf{Relation-attention weights}. We believe that learning weights from both a global feature and local features is more reliable.
The self-attention weights are learned with individual frame features and non-linear mapping, which are rather coarse. Since $f'_v$ inherently contains the contents of the whole video, the attention weights can be further refined by modeling the relation between frame features and this global representation $f'_v$.

Inspired by the relation-Net in low-shot learning~\cite{yang2018learning}, we use the sample concatenation and another FC layer to estimate new relation-attention weights for frame features. The relation-attention weight of the $i$-th frame is formulated as,
\begin{equation}
\beta_{i} = \sigma([f_{i}:f'_v]^T\mathbf{q}^1),
\end{equation}
where $\mathbf{q}^1$ is the parameter of FC, $\sigma$ denotes the sigmoid function. 

Finally, with self-attention and relation-attention weights, our FAN aggregates all the frame features into a new compact feature as,
\begin{equation}
\label{eq:alphabeta}
f_v = \frac {\sum_{i=0}^{n}\alpha_{i}\beta_{i}[f_{i}:f'_v]}{\sum_{i=0}^{n}\alpha_{i}\beta_{i}}.
\end{equation}

\section{Experiments}
\label{sec:experiments}
\subsection{Datasets and Implementation Details}
\textbf{CK+}~\cite{lucey2010extended} contains 593 video sequences from 123 subjects. Among these videos, 327 sequences from 118 subjects are labeled with seven basic expression labels, i.e. anger, contempt, disgust, fear, happiness, sadness, and surprise. Since CK+ does not provide training/testing splits, most of the algorithms evaluated on this database with 10-fold person-independence cross-validation experiments. We constructed 10 subsets by sampling ID in ascending order with a step size of 10 as in several previous works~\cite{liu2014learning,kuo2018compact}, and report the overall accuracy over 10 folds.

\textbf{AFEW 8.0}~\cite{dhall2018emotiw} served as an evaluation platform for the annual EmotiW since 2013. Seven emotion labels are included in AFEW, i.e. anger, disgust, fear, happiness, sadness, surprise and neutral. AFEW contains video clips collected from different movies and TV serials with spontaneous expressions, various head poses, occlusions, and illuminations. AFEW 8.0 is divided into three splits: Train (773 samples), Val (383 samples) and Test (653 samples), which ensures data in the three sets belong to mutually exclusive movies and actors. Since the test split is not publicly available, we train our model on training split and report results on validation split.

\textbf{Implementation details}.
We preprocess video frames by face detection and alignment in the Dlib toolbox
We extend the face bounding box with a ratio of 25\% and then resize the cropped faces to scale of 224$\times$224.
We implement our method by the Pytorch toolbox.
By default, for feature embedding, we use the ResNet18 which is pre-trained on MS-Celeb-1M~\cite{guo2016ms} face recognition dataset and FER\_Plus expression dataset \cite{BarsoumICMI2016}.
For training, on both CK+ and AFEW 8.0, we set a batch to have 48 instances with $K$ frames in each instance. For frame sampling in a video, we first split the video into $K$ segments and then randomly select one frame from each segment. By default, we set $K$ to 3. We use the SGD method for optimization with a momentum of 0.9 and a weight decay of $10^{-4}$.
On CK+, we initialize the learning rate (\textit{lr}) to $0.1$, and modify it to 0.02 at 30 epochs, and stop training after 60 epochs. 
On AFEW 8.0, we initialize the \textit{lr} to 4e-6, and modify it to 8e-7 at 60 epochs and 1.6e-7 at 120 epochs, and stop training after 180 epochs. 

\begin{table}[tp]
\center
\caption{Evaluation of our FAN with a comparison to state-of-the-art methods on CK+ database. Note that only those methods evaluated with 7 classes are included.}
\begin{tabular}{p{60pt}p{60pt}p{60pt}p{20pt}}
 \toprule
Method& Training data & Test data& Acc.\\
 \midrule
\small{ST network~\cite{zhang2017facial}} & \tabincell{l}{S: the last frame\\T: all frames} & \tabincell{l}{S: the last frame\\T: all frames} & 98.50 \\
 \midrule
DTAGN~\cite{Jung2015Joint} & Fixed length & Fixed length  & 97.25 \\ 
 \midrule
CNN+Island loss~\cite{cai2018island} & \tabincell{l}{The last three \\frames and the \\first frame} & \tabincell{l}{The last three \\frames and the \\first frame} & 94.35 \\
 \midrule
LOMo~\cite{sikka2016lomo} & All frames & All frames  & 92.00 \\ 
  \midrule
\tabincell{l}{Score fusion\\ (baseline) }&All frames & All frames  & 94.80 \\
FAN(w/o Relation-attention) & All frames & All frames & \textbf{99.08} \\
FAN & All frames & All frames & \textbf{99.69} \\
 \bottomrule
\end{tabular}
\label{tab:ck+}
\end{table}
\subsection{Evaluation on CK+}
We evaluate our FAN on CK+ with comparisons to several state-of-the-art methods in Table \ref{tab:ck+}. 
On CK+, due to the fact that the videos show a shift from a neutral facial expression to the peak expression, most of the methods conduct data selection manually. Zhang et al~\cite{zhang2017facial} propose to combine a spatial CNN model and a temporal network, where the spatial CNN model only uses the last peak frame. Jung et al~\cite{Jung2015Joint} select a fixed length sequence for each video with a lipreading method~\cite{ZihengZhou:2011}, and jointly fine-tune a deep temporal appearance-geometry network. Cai et al~\cite{cai2018island} select the last three frames and the first frame for each video, and train CNN models with a new Island loss function.
We argue that\textit{ manual data selection is an ad-hoc operation on CK+ and it is impractical since we can not know which is the peak frame beforeahead}. Sikka et al~\cite{sikka2016lomo} use all frames with a new latent ordinal model which extracts CNN/LBP/SIFT features for sub-event detection and uses multi-instance SVM for expression classification.
Our baseline method uses ResNet18 to generate scores for individual frame and applies score fusion (summation) for all frames. It achieves 94.8\% which is 2.8\% better than \cite{sikka2016lomo}. Our proposed FAN with only self-attention gets 99.08\% which significantly boosts the baseline by 4.28\%. Adding relation-attention weights further improves the accuracy to 99.69\% which sets up a new state of the art on CK+.
\subsection{Evaluation on  AFEW 8.0}
From the view of performance, AFEW is one of the most challenging videos FER dataset. The EmotiW challenge shares the same data from 2016 to 2018. 
Table \ref{tab:afew} presents the evaluation of our FAN on AFEW with comparisons to recent state-of-the-art methods. For a fair comparison, we only list these results obtained by the best single models in previous works. From the last three rows of Table \ref{tab:afew}, our proposed FAN improves the baseline by 2.36\%. 
Both \cite{fan2016video} and \cite{Vielzeuf2017Temporal} use VGGFace backbone and a recurrent model with long-short-term memory units. These methods aim to capture temporal dynamic information for videos.
Most of the methods focus on improving static face based CNN models and combine scores for video-level FER. Both \cite{yao2016holonet} and \cite{Hu2017Learning} input two LBP maps and a gray image for CNN models. Deeply-supervised networks are used in \cite{Hu2017Learning} and \cite{fan2018video}, which add supervision on intermediate layers.
For static methods, \cite{liu2018multi} gets slightly better performance than ours. However, \cite{liu2018multi} uses DenseNet-161 and pretrains it on both large-scale face datasets and their own Situ emotion video dataset. Additionally, \cite{liu2018multi} applies complicated post-processing which extracts frame features and compute their mean vector, max-pooled vector, and standard deviation vector. These vectors are then concatenated and finally fed into an SVM classifier.
Overall, our FAN improves the baseline significantly and achieves performance comparable to that of the best previous single model.

\begin{table}[tp]
\center

\caption{Evaluation of our FAN with a comparison to state-of-the-art methods on AFEW 8.0 database. It is worth noting that we only compare to the best \textit{single} models of previous works.}
\resizebox{\linewidth}{!}{
\begin{tabular}{ccc}
 \toprule
Method& Model type & Accuracy\\
 \midrule
CNN-RNN (2016)~\cite{fan2016video} & Dynamic & 45.43 \\ 
VGGFace + Undirectional LSTM (2017)~\cite{Vielzeuf2017Temporal} & Dynamic & 48.60 \\
HoloNet (2016)~\cite{yao2016holonet} & Static & 44.57 \\ 
DSN-HoloNet (2017)~\cite{Hu2017Learning} & Static & 46.47 \\
DenseNet-161 (2018)~\cite{liu2018multi} & Static & \textbf{51.44} \\ 
DSN-VGGFace (2018)~\cite{fan2018video} & Static & 48.04 \\
 \midrule
Score fusion (baseline)  & Static & 48.82 \\
FAN w/o Relation-attention& Static & \textbf{50.92} \\
FAN & Static & \textbf{51.18} \\
 \bottomrule
\end{tabular}}
\label{tab:afew}
\end{table}

\subsection{Visualization and Hyper-parameters}
\begin{figure}[tp]
\centering
\includegraphics[width=0.498\linewidth]{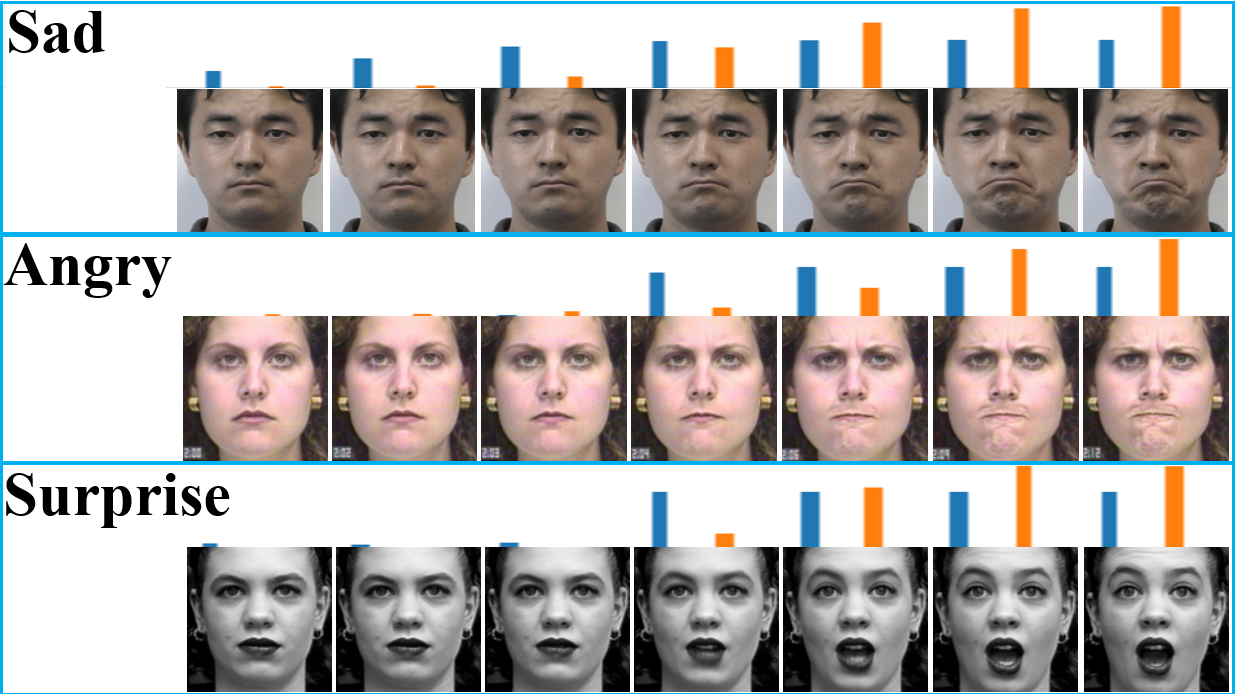}
\includegraphics[width=0.492\linewidth, height=0.28\linewidth]{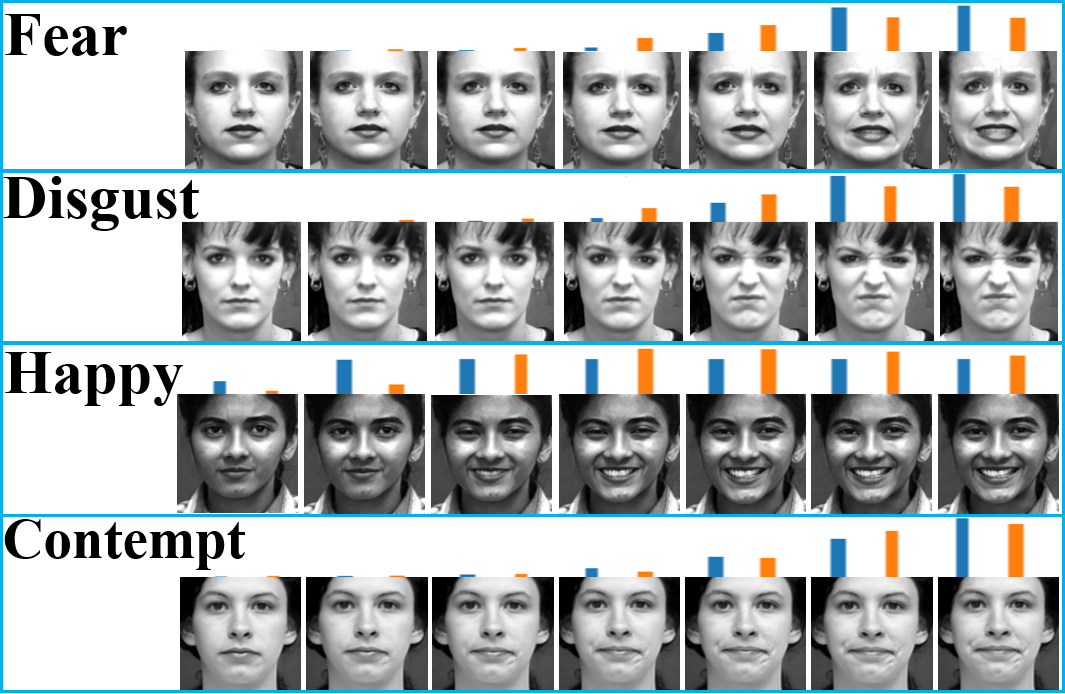}
\caption{Visualization of the self-attention weights (blue bar) and the final weights of FAN (orange bar) on CK+ dataset. }
\label{fig_Visulization}
\end{figure}

To better understand the self-attention and relation-attention modules in our FAN, we visualize the attention weights in Figure \ref{fig_Visulization}. Figure \ref{fig_Visulization} shows one sequence for each category with blue and orange weight bars, where blue bars represent the self-attention weights (i.e. $\alpha$ in Eq. (\ref{eq:alpha})) of our FAN w/o relation-attention and orange bars the final weights (i.e. $\alpha\beta$ in Eq. (\ref{eq:alphabeta})) of  our FAN.
In total, both kinds of weights can reflect the importance of frames.
Comparing the blue and orange bars, we find that the final weights of our FAN can always assign higher weights to the more obvious face frames, while self-attention module could assign high weights on some obscure face frames, see the 1st, 2th, and 3rd rows of Figure \ref{fig_Visulization} (left). This explicitly explains why adding relation-attention boost performance. 

\begin{figure}[tp]
\centering
\includegraphics[width=1\linewidth]{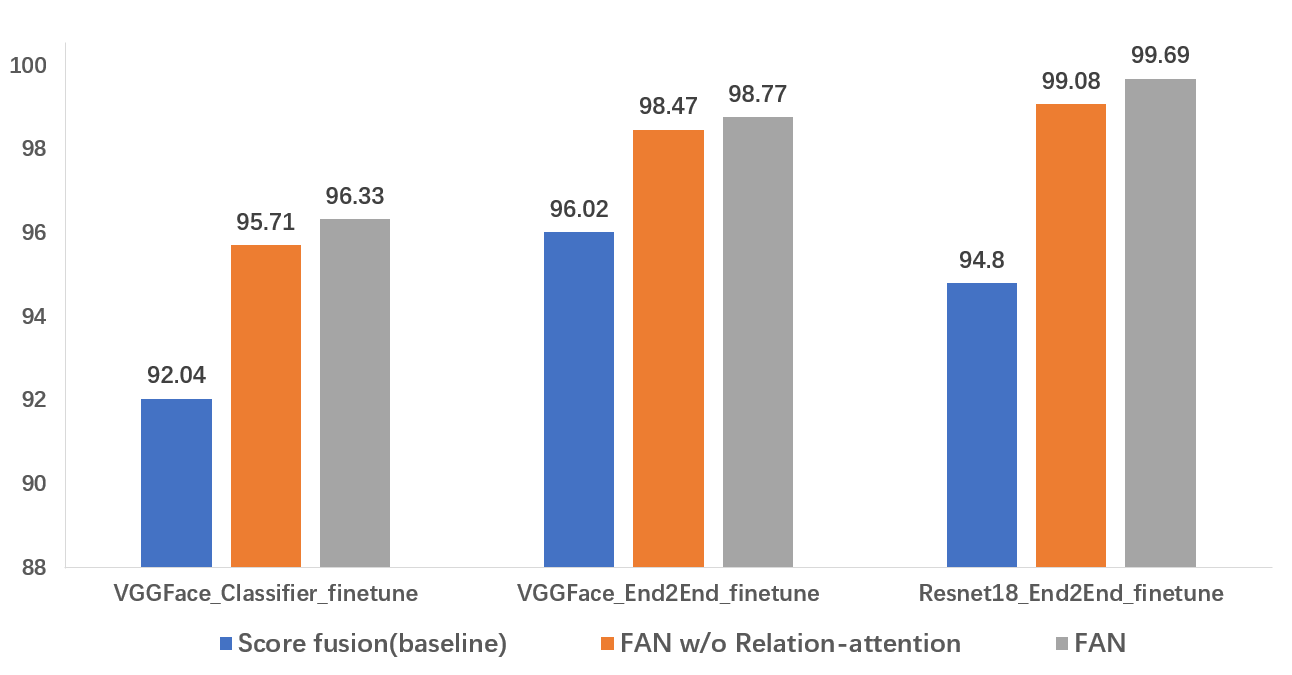}
\caption{Evaluation of backbone CNN models and training strategies on CK+.}
\label{fig_VGGResNet}
\end{figure}
\textbf{Evaluation of Hyper-parameters}.
We evaluate two hyper-parameters of our FAN on CK+, i.e. backbone CNN networks and the parameter $K$ mentioned in implementation details, to validate the robustness of our method. 
For the parameter $K$,  besides the default value, we try several other values, i.e. \{2, 5, 8\}, and find the performance is not sensitive to $K$. Specifically, our FAN obtains 99.39\% with $K$=\{2, 5\}. and gets 99.69\% with $K$=8. Since the default value, $K$=3 gets 99.69\%, we use this default setting in the remainder of this paper.

For the backbone CNN model evaluation, we try the VGGFace model which is widely-used in previous works. Similarly, we also pretrain the VGGFace model on the FERPlus dataset. Since \cite{Knyazev2017Convolutional} shows that it is better to freeze all the feature learning layers after pretrained on FERPlus for VGGFace model, we also conduct the same experiment on CK+ with VGGFace.
Figure \ref{fig_VGGResNet} shows the default comparisons with different backbone CNN models.
On CK+, compared with freezing all the feature layers for VGGFace, it gets better results with fine-tuning all layers which may be explained by the domain discrepancy between FERPlus and CK+. Overall, the results are significantly improved by self-attention weights and further improved by the relation-attention weights.

\section{Conclusion}
\label{sec:conclusion}
We propose Frame Attention Networks for video-based facial expression recognition. The FAN contains a self-attention module and a relation-attention module. The experiments on CK+ and AFEW show that our FAN with only self-attention improves the baseline significantly and adding relation-attention further boosts performance. With a visualization on CK+, we demonstrate that our FAN can automatically capture the importance of frames. Our single model achieves performance on par with that of state-of-the-art methods on AFEW and obtains state-of-the-art results on CK+.



\small
\bibliographystyle{IEEEbib}
\bibliography{strings,refs}

\end{document}